\documentclass{article}

\usepackage{arxiv}

\usepackage{booktabs}						

\usepackage{amsfonts}						
\usepackage{graphicx}

\usepackage{placeins}
\usepackage{wrapfig}

\usepackage[numbers,compress,sort]{natbib}	

\usepackage{multirow}

\usepackage{amsmath}
\usepackage{authblk}
\usepackage{hyperref}

\usepackage{fontawesome}

\title{Multi-Class and Multi-Task Strategies\\ for Neural Directed Link Prediction}
\author[1,2]{Claudio Moroni \href{mailto:claudio.moroni@dedagroup.it}{\faEnvelopeO}}
\author[2]{Claudio Borile}
\author[2]{Carolina Mattsson}
\author[2,3]{Michele Starnini}
\author[2]{André Panisson}

\affil[1]{Dedagroup, Torino, Italy}
\affil[2]{CENTAI Institute, Torino, Italy}
\affil[3]{Department of Engineering, Universitat Pompeu Fabra, Barcelona, 08018, Spain}

\begin{document}

\maketitle

\begin{abstract}
Link Prediction is a foundational task in Graph Representation Learning, supporting applications like link recommendation, knowledge graph completion and graph generation. Graph Neural Networks have shown the most promising results in this domain and are currently the de facto standard approach to learning from graph data.
However, a key distinction exists between Undirected and Directed Link Prediction: the former just predicts the existence of an edge, while the latter must also account for edge directionality and bidirectionality. This translates to Directed Link Prediction (DLP) having three sub-tasks, each defined by how training, validation and test sets are structured.
Most research on DLP overlooks this trichotomy, focusing solely on the "existence" sub-task, where training and test sets are random, uncorrelated samples of positive and negative directed edges. Even in the works that recognize the aforementioned trichotomy, models fail to perform well across all three sub-tasks.\\
In this study, we experimentally demonstrate that training Neural DLP (NDLP) models only on the existence sub-task, using methods adapted from Neural Undirected Link Prediction, results in parameter configurations that fail to capture directionality and bidirectionality, even after rebalancing edge classes. 
To address this, we propose three strategies that handle the three tasks simultaneously. Our first strategy, the Multi-Class Framework for Neural Directed Link Prediction (MC-NDLP) maps NDLP to a Multi-Class training objective. The second and third approaches adopt a Multi-Task perspective, either with a Multi-Objective (MO-DLP) or a Scalarized (S-DLP) strategy.
Our results show that these methods outperform traditional approaches across multiple datasets and models, achieving equivalent or superior performance in addressing the three DLP sub-tasks.

\end{abstract}

\section{Introduction} \label{sec-introduction}

Graphs are a natural way to represent complex systems. Examples include social networks, financial transaction networks, power grids, and neuronal connectivity~\cite{albert2002statistical, OGB}.
These systems can be modeled using different types of graphs, ranging from simple networks to more sophisticated structures like Knowledge Graphs \cite{hogan2021knowledge}, Dynamic Graphs \cite {Holme2019}, or Bipartite Graphs for recommender systems~\cite{Li2009}.
Given the broad applicability of graphs, Representation Learning on graph-like data structures has become essential, with core tasks including node classification, link prediction and graph classification.

In this work, we focus on Link Prediction \cite{LU20111150}, particularly its Directed variant. Recently, Graph Neural Networks
(GNN)-based models, such as Graph Autoencoders, have been devised to address Link Prediction tasks \cite{kipf2016variational, GASN, SEAL, Cai_Ji_2020, MAGNN, CPAGCN, Kollias2022DirectedGA, DGLP, 2S-SDGCN, MVGAE, Salha2019, zhang2021magnet, ClusterLP}, establishing the field of Neural Link Prediction. These models have several important applications, including completing knowledge graphs~\cite{TransE} 
, serving as baseline for deep graph generation \cite{kipf2016variational, GraphVAE} and pre-processing transaction networks~\cite{LIN2022_EthereumTracking}. The recent literature primarily focuses on undirected applications~\cite{KUMAR2020124289,Qin2022,Wu2022}, with few studies mentioning directed cases~\cite{Arrar2023}. Direction can be core to the application itself, in some domains. With citation graphs, for instance, citing and being cited have substantively different meanings. Moreover, incorporating edge direction has been shown to improve learning for node classification across different types of graphs~\cite{rossi_edge_2024}. Even so, the complexity of Directed Link Prediction (DLP) is often overlooked and it has been argued that this limits progress in NDLP~\cite{Salha2019}.

Neural Directed Link Prediction (NDLP) requires a model that is capable of representing edge direction and a training strategy that effectively learns directionality. Not all GNNs can represent edge direction. Graph Autoencoders, for example, often use decoder implementations where probabilities for edges $(u,v)$ and $(v,u)$ are the same by design~\cite{Salha2019}. We refer to these models as NDLP-incapable. But even NDLP-capable models can fail to learn edge directionality when the training strategy is borrowed from Neural Undirected Link Prediction (NULP). 
Typically, NULP models are trained and evaluated on random subsets of positive and negative undirected edges~\cite{kipf2016variational, GASN, SEAL, Cai_Ji_2020, MAGNN, CPAGCN, ClusterLP}. Now, on a sparse directed graph, it is statistically unlikely that a random subset of negative directed edges would include the reverse of a randomly sampled positive directed edge. This allows models to ignore edge direction without incurring a penalty. Indeed, using the NULP approach to training and evaluating NDLP models~\cite{Kollias2022DirectedGA, DGLP, 2S-SDGCN, MVGAE} can lead to NDLP-incapable models performing deceptively well.

Training and evaluation for NDLP is more complex. Three sub-tasks have been devised in the recent literature to comprehensively evaluate distinct aspects of DLP~\cite{Salha2019, zhang2021magnet, ClusterLP, ADGE}. The ``General'' DLP task is the classic adaptation of the approach used in ULP to the directed case. This is then complemented with two other binary classification sub-tasks designed to test a model's ability to distinguish edge directions. Namely, the ``Directional'' and ``Bidirectional'' sub-tasks. Prior work has shown that NDLP-capable models can learn to perform well on each of the three sub-tasks~\cite{Salha2019, zhang2021magnet, ClusterLP, ADGE} and that there is a trade-off among them~\cite{zhang2021magnet}. However, prior approaches do not consider training strategies that can handle these three sub-tasks simultaneously.

Here we propose three learning strategies to improve performance across DLP sub-tasks, simultaneously. 
The first strategy, \textit{Multi-Class Directed Link Prediction} (MC-DLP), maps DLP to a four-class classification task.
This framework distinguishes between unidirectional positives, unidirectional negatives, bidirectional positives, and bidirectional negatives, ensuring balanced contributions to the training loss.
The other two approaches recognize the Multi-Task nature of DLP, simultaneously training on simultaneously constructed \textit{General DLP}, \textit{Directional} and \textit{Bidirectional} training sets.
Drawing from the literature on Multi-Objective \cite{MGDA} and Scalarization \cite{RevisitingScalarization} methods, we propose
\textit{Multi-Objective Directed Link Prediction} (MO-DLP) and \textit{Scalarization-based Directed Link Prediction} (S-DLP) strategies to handle these tasks more effectively. Each of our three training strategies incentivize NDLP-capable models to perform well across different aspects of DLP, and our results show that better training strategies can be just as important as better models in advancing the state of Neural Directed Link Prediction.

The remainder of this paper is organized as follows: Section \ref{sec-back} covers the background concepts and related work, highlighting their relevance to our approach. In Section \ref{sec-methods}, we detail the proposed multi-class and multi-task strategies. Section \ref{sec-experiments} outlines the experimental setup, describes the datasets and presents a performance comparison  of various models across all strategies and datasets. Finally, Section \ref{sec-conclusions} provides concluding remarks.

\section{Background and Related Work} \label{sec-back}

In this section, we introduce the notation and review key concepts and prior research relevant to Neural Directed Link Prediction (NDLP). We briefly discuss foundational work in Graph Neural Networks and their applications in undirected link prediction, then go on to examine approaches for incorporating directionality. This serves to highlight certain limitations and the need for more comprehensive training strategies.

\subsection{Notation}
\label{sec:notation}
Given a directed graph $G = (V,E)$ with $N = |V|$ nodes where $E \subseteq V \times V$, and given $u,v \in V$ we say that:

\begin{itemize}
    \item $(u,v)$ is \textit{negative bidirectional} $\iff$ $(u,v) \notin E \land (v,u) \notin E$;
    \item $(u,v)$ is \textit{negative unidirectional} $\iff$ $(u,v) \notin E \land (v,u) \in E$;
    \item $(u,v)$ is \textit{positive unidirectional} $\iff$ $(u,v) \in E \land (v,u) \notin E$;
    \item $(u,v)$ is \textit{positive bidirectional} $\iff$ $(u,v) \in E \land (v,u) \in E$;
\end{itemize}

Moreover, we denote $A \in \{0,1\}^{N \times N}$ as $G$'s adjacency matrix, and $X \in \mathbb{R}^{N \times F}$ as node features.

\subsection{Graph Neural Networks} \label{sec-GNN}

Graph Neural Networks (GNNs) are currently the de facto standard approach for directed link prediction.
Message Passing Neural Networks (MPNNs) \cite{Gilmer2017,GraphSAGE,veličković2018graph,xu2018how} is the most general framework for GNNs, and other frameworks such as  Spectral Graph Convolutional Neural Networks (SGCNNs)~\cite{Defferrard2016,kipf2017semisupervised} can be mapped to the MPNN framework.

Applications concern pharmaceutical industry, material science \cite{Reiser2022}, time series \cite{jin2023survey}, anti-money laundering \cite{johannessen2023finding,weber2019antimoney,egressy2024provably,altman2023realistic}. In the remainder of this section, we describe the general MPNN paradigm, current state-of-the-art deep learning technique for Graph Representation Learning. 

MPNNs elaborate successive hidden representations for each node $v$ aggregating messages from its neighbors.

Given $z^{(k)}_v$ as the $k$-th layer embedding of node $v$ (with $z^{(0)}_v$ equalling $v$'s feature vector $x_v$), GNNs compute the next layer embedding $z^{(k+1)}_v$ through the following relation:
\begin{equation}
    z^{(k+1)}_v = f^{(k)}\left(m_s^k(z^{(k)}_v), A^{(k)}\left(\left\{m_n^{(k)}(z^{(k)}_u) | u \in N(v)\right\}\right)\right)
\end{equation}
Where $f^{(k)}, A^{(k)}, m_s^k \text{ and } m_n^{(k)}$ are, respectively, the layer-specific update function, the aggregation function, the self-information and message functions.
$N(v)$ indicates the neighborhood of node $v$.

If we consider $K$ layers in total, the last embedding of each node $z_v\equiv z^{(K)}_v$ can be used for downstream tasks such as node classification, link prediction, and graph classification. For link prediction, a \textit{decoder} function takes as input the representation of two nodes, $z_u, z_v$ and outputs a normalized score representing the probability of an edge $(u, v)$ being present. For the undirected case, the decoder function can be the scalar product followed by a sigmoid function, $DEC(z_u, z_v)=\sigma(z_u\cdot z_v)$, or a neural network such as a MLP. The scalar product, being symmetric in $u$ and $v$, is an example of a decoder unable to learn directionality.

\begin{figure}
    \centering
    \includegraphics[width=0.7\linewidth]{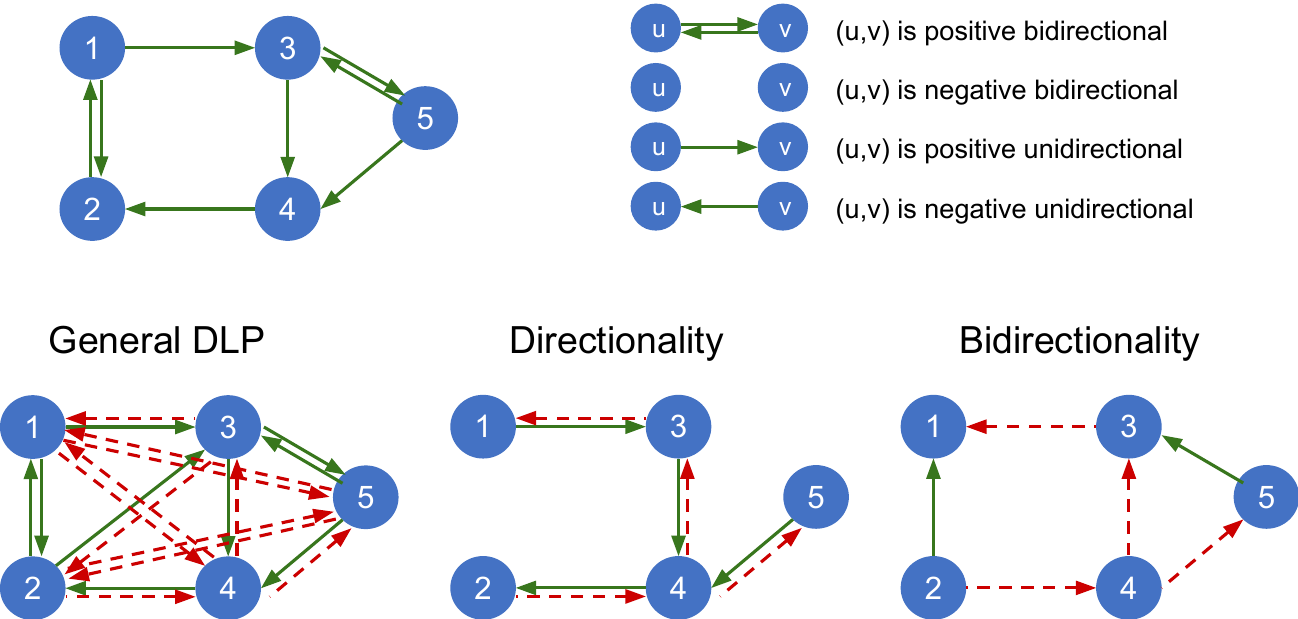}
    \caption{
    From a graph $G$ (top panel), for each task definition, green edges are the positive class and red edges are the negative class.
    In \textit{General DLP}, any absent directed edge can be selected as negative. For \textit{Directionality} prediction, unidirectional edges are positive, with their inverses as negatives. For \textit{Bidirectionality} prediction, one direction of bidirectional edges is positive, and the reverse of unidirectional edges is negative.
    }
    \label{fig:training_sets}
    \vspace{-0.5cm}
\end{figure}

\subsection{Neural Directed Link Prediction} \label{sec-related}

This work considers three sub-tasks for directed link prediction, as presented in \cite{Salha2019}. Figure~\ref{fig:training_sets} shows a representation of the positive and negative classes that define the \emph{General Directed Link Prediction (General DLP)}, \emph{Biased Directional Negative Samples Link Prediction (Directional)}, and \emph{Bidirectionality Prediction (Bidirectional)} tasks. These classes are the basis for selecting the (directed) edges used in the testing and evaluation of NDLP-capable models on DLP in the most relevant recent literature.

We focus on NDLP-capable models that use Graph Autoencoder techniques. \cite{Kollias2022DirectedGA} develops an extension of the Weisfeiler-Lehman kernel \cite{WLKernel} which is then used as a basis to define a source/target-like \cite{Salha2019} graph autoencoder for directed graphs; however, the model is only tested on the \textit{General DLP} sub-task. \cite{zhang2021magnet} defines a GCN for directed graphs where the aggregation is performed by a complex, hermitian laplacian. This model is tested across all three tasks, although for each of them, a different parameter set is inferred. To the best of our knowledge, \cite{Salha2019} devises the earliest GNN-based autoecoders for NDLP, and is among the first to highlight the intrinsic differences between NDLP and NULP; Their work is also responsible for one of the earliest usage of the \textit{General DLP, Directional} and \textit{Bidirectional} sub-tasks in a neural setting. Similarly to \cite{zhang2021magnet}, the authors of \cite{Salha2019} do not find one parameter set for all three tasks for each model. \cite{ClusterLP} endows cluster information in node embeddings, but, similarly to \cite{Salha2019}, trains and tests over the \textit{Directional} and \textit{Bidirectional} tasks using two different training graphs. \cite{ADGE} produces unsupervised source/target node embeddings by adversarially training a neural pair of generator and discriminator on the graph topology; the final model performs simultaneously well on the \textit{General DLP} and \textit{Directional} task, but it is not evaluated on the \textit{Bidirectional} task. 

Although we will experiment with different models, it is not our aim to sponsor anyone in particular. Rather, we propose learning strategies that can used with \textit{any} NDLP-capable model to encourage encoding directionality and strike a better balance among the performances on sub-tasks of NDLP. Therefore our work is much more in the spirit of \cite{MaskGAE}, a representation learning framework that argues in favor of masking compared to full-graph training. Unfortunately, \cite{MaskGAE} has been developed and tested on undirected graphs only, so its extension to directed graphs could be a future research avenue.

\section{Strategies for Neural Directed Link Prediction} \label{sec-methods}
In the previous section, we discussed that NDLP has recently been described as three distinct sub-tasks: General DLP, Directional, and Bidirectional. Simultaneously addressing these requires more than adapting the classic approach to (undirected) link prediction. In this section, we formalize two training strategies that are in principle applicable to every NDLP-capable encoder-based MPNN and encourage such models to learn to encode directionality.

\subsection{Multi-Class Strategies for Neural Directed Link Prediction} 
\label{sec-MCNDLP}

As anticipated in Section~\ref{sec-introduction}, we assume that the reason why prior works are generally not able to infer a single model that performs well on all three sub-tasks is related to an unaddressed imbalance between unidirectional and bidirectional edges' contributions to the training loss. We note that this imbalance should be dealt with without compromising the positive/negative edges' reweighting. Therefore we propose to simultaneously balance positives vs negatives and unidirectional vs bidirectional edges using the following Multi-Class Neural Directed Link Prediction (\textit{MC-NDLP}) strategy. 

Given a GNN model that computes $d_K$-dimensional embeddings $z_v,\ \forall v \in V$, we may compute logits for each of the four classes listed in Section \ref{sec:notation} by applying an MLP to the concatenation of the embeddings. The MLP must take $2d_K$ input dimensions and output $4$ logits, and can be arbitrarily deep:
\begin{equation}
    [\hat{l}^{nb}_{uv}, \hat{l}^{nu}_{uv}, \hat{l}^{pu}_{uv}, \hat{l}^{pb}_{uv}] = \text{MLP}(z_u || z_v),
\end{equation}
where $\hat{l}^{m}_{uv}, m \in \{nb, nu, pu, pb\}$ denote the model's output logits for the edge $(u, v)$ being negative bidirectional, negative unidirectional, positive unidirectional, or positive bidirectional as defined in \S\ref{sec:notation}.

Notably, \textit{MC-NDLP} is also compatible with any graph autoencoder that makes use of specific decoders which output only one logit $\hat{l}_{uv}$, that is, the model output for the presence of a directed edge $(u, v)$ \cite{Salha2019,Kollias2022DirectedGA}. We can turn the standard binary classification task for NDLP into a 4-class classification task by transforming the output logit into a probability via e.g., a sigmoid

\begin{equation}
\hat{p}_{uv} = \sigma(\hat{l}_{uv})
\end{equation}

and defining:

\begin{equation}
\begin{split}
    [\hat{p}^{nb}_{uv}, \hat{p}^{nu}_{uv}, \hat{p}^{pu}_{uv}, \hat{p}^{pb}_{uv}] = [(1 - \hat{p}_{uv})(1 - \hat{p}_{vu}), \\  (1 - \hat{p}_{uv})\hat{p}_{vu}, \\ \hat{p}_{uv}(1 - \hat{p}_{vu}), \\ \hat{p}_{uv}\hat{p}_{vu} ].
\end{split}
\label{eq-multiclass_extension}
\end{equation}

We note that Eq.~\ref{eq-multiclass_extension} assumes statistical independence between $\hat{p}_{uv}$ and $\hat{p}_{vu}$, which are both conditioned on $A$ and $X$. Infact, most autoencoders model $\hat{p}_{uv} = \hat{P}(e_{uv} | A,X)$ where $e_{uv} = 1 \iff (u,v) \in E$ otherwise it equals $0$ \cite{kipf2016variational}, and Eq.~\ref{eq-multiclass_extension} naturally extends these univariate autoencoders. Please refer to Appendix \ref{sec:stat-indep} for more details.

We can then define a weighted Multi-Class cross-entropy loss function:
\begin{equation}
    \mathcal{L}_{\textit{MC-NDLP}}(\Theta) = -\sum\limits_{c\in C}\sum\limits_{uv\in T}w_{y_{uv}}\mathbb{I}(y_{uv}=c)\log(\hat{p}^{y_{uv}}_{uv}),
\end{equation}
where $C=\{nb, nu, pu, pb\}$, $T$ is the \textit{General DLP} training set (see \S\ref{sec-split}), $\mathbb{I}$ is the indicator function and $y_{uv}\in C$ is the ground-truth class of the edge $(u, v)$, and $\hat{p}^{y_{uv}}_{uv}$ is the model's output probability of edge $(u,v)$ belonging to the ground-truth class $y_{uv}$. The class weight is defined as 
\begin{equation}
    w_{y_{uv}} = \frac{n_{x}}{n_{y_{uv}}},
\end{equation}
where $n_x$ represents the number of samples in class $x$, where $x$ is the most numerous class (usually \textit{nb}), and  $n_{y_{uv}}$ is the number of samples in class $y_{uv}$. As discussed in \S\ref{sec-introduction}, this class reweighting mitigates the statistical imbalance between all four classes defined in \S\ref{sec:notation}.

\subsection{Multi-Task Strategies for Neural Directed Link Prediction} \label{sec-MTNDLP}
Multi-Task Learning (MTL) refers to scenarios where more than one objective function must be simultaneously optimized. It is more challenging compared to single-task learning, due to the various objectives having no a priori relative importance and generally competing against each other. To simultaneously exploit the \textit{General DLP, Directional} and \textit{Bidirectional} training sources of information, we devise a multi-task objective over the three sub-tasks, defined by the binary cross-entropy loss functions on \textit{General DLP} $\mathcal{L}_G$, \textit{Directional} $\mathcal{L}_D$ and \textit{Bidirectional} $\mathcal{L}_B$.

Multi-Task learning is usually carried out in two ways:

\begin{itemize}
    \item \textit{Scalarization}: it prescribes to sum and weight the losses to reduce the optimization to the single-objective case:$$\mathcal{L} = \alpha_G \mathcal{L}_G + \alpha_D \mathcal{L}_D + \alpha_B \mathcal{L}_B$$ The coefficients $\alpha_i,\ i=G,D,B$, can be either learned or heuristically set. Despite its simplicity, heuristic implementations of Scalarization have been proven to achieve competitive performance in real use cases \cite{RevisitingScalarization}. In this work, we set them to the validation losses (normalized between 0 and 1) of the previous epoch, to favor generalization. We name this approach \textit{S-NDLP};

    \item \textit{Multi-Objective}: it consists in finding a parameter update rule that ensures that all losses are diminished (or left unchanged) at each optimization step. Given a model $f_{\Theta}$ and the objectives $\{\mathcal{L}_i(\Theta)\}_{i = 1}^{L}$ to be simultaneously optimized, we say that $\Theta_1$ \textit{dominates} $\Theta_2 \iff \mathcal{L}_i(\Theta_1) \leq \mathcal{L}_i(\Theta_2) \qquad \forall i \in 1,...,L$. Therefore, a solution $\Theta^*$ is \textit{Pareto-optimal} if it is not dominated by any other solution. The set of non-dominated solutions is called the Pareto set $\mathcal{P}$. While many Gradient-based Multi-Objective optimization algorithms with guaranteed convergence on the Pareto set have been developed, we focus for simplicity on one of the first, MGDA~\cite{MGDA}. This algorithm is based on the observation that given the gradients associated with the individual losses, the opposite of their shortest convex linear combination points in the direction where all losses either remain constant or diminish. We name this approach \textit{MO-NDLP}.
\end{itemize}

\subsection{Simultaneous Splits} \label{sec-split}

Train, test, and validation sets are constructed from the positive and negative classes associated to each of the three DLP sub-tasks (see Fig. \ref{fig:training_sets}). In~\cite{Salha2019}, edges are randomly sampled from the positive classes to construct the validation and test sets separately for each sub-task. Since models are separately trained on each sub-task, the training sets can be defined as what remains when the positive samples for that sub-task are reserved. In our case, the models must be trained over the graph that remains when \textit{all} the reserved edges are removed. In order to preserve as many edges as possible for training, sampling is done together. Specifically, for each dataset, a random $10\%$ and $5\%$ of all unidirectional edges are reserved for testing and validation, respectively. Moreover, a random $30\%$ and $15\%$ of (one direction of) all bidirectional edges are reserved for testing and validation, respectively.

Validation and test sets for the three sub-tasks are constructed using the reserved edges. All the sampled edges (unidirectional and bidirectional) are used as positives for the \textit{General DLP} task, complemented with an equal number of randomly sampled absent directed edges as negatives. The sampled unidirectional edges are used as positives for the \textit{Directional} task, complemented with their reverses as negatives. The randomly sampled bidirectional edges are used for the \textit{Bidirectional} task, complemented with an equal number of the reverses of unidirectional edges randomly sampled from the remaining graph as negatives. To perform well on this task, the model must distinguish between edges whose reciprocal exists from those whose reciprocal does not.

Multi-class learning requires a single training set. As in the \textit{General DLP} task of \cite{Salha2019}, and in the classic formulation of directed link prediction, the training set is constructed out of all remaining directed edges and all the absent edges in the incomplete training graph. In our case, these edges are sorted into the four classes defined in \S\ref{sec:notation} for multi-class classification as described in \S\ref{sec-MCNDLP}. 

Multi-task learning requires training sets for each sub-task. Here, we take the opportunity to construct our own versions of the \textit{training} sets for the \textit{Directional} and \textit{Bidirectional} tasks. This is done so that the training set has similar edge statistics compared to their respective validation and test sets. In particular: for the \textit{Directional} task, the training set is composed of all the remaining unidirectional edges (as positives) and their reverses (as negatives). This is similar to the training set for the \textit{General DLP} task, but with the bidirectional edges removed. Finally, for the \textit{Bidirectional} task, one direction of all the remaining bidirectional edges are used, as positives, together with an equal amount of the reverses of unidirectional edges randomly sampled from the remaining graph, as negatives. 

These modifications to the sampling described in \cite{Salha2019} make simultaneous training possible while ensuring no overlap between train and test data. Validation and test sets for the three sub-tasks are constructed using the reserved edges, and training sets are constructed using the remaining edges.

\section{Experiments} \label{sec-experiments}

In this section, we evaluate the effectiveness of our proposed strategies through comparative experiments using well-known datasets and DLP models. We aim to demonstrate the performance improvements of our approaches across multiple tasks and models, highlighting their ability to handle the challenges of edge directionality and directionality.
All code and results are publicly available at \url{https://github.com/ClaudMor/Multi_Task_Multi_Class_Neural_Directed_Link_Prediction}.

\subsection{Datasets}

Our experiments are conducted on three publicly available datasets, each of which is a directed graph. As in \cite{Salha2019}, we consider two small citation networks (Cora and CiteSeer) and a larger hyperlink network (Google).

\begin{table}[h!]

\caption{Network statistics of the datasets used, computed using \texttt{networkx} \cite{hagberg_exploring_2008}. }

\label{datasets}
\centering
\resizebox{0.8\linewidth}{!}{\begin{tabular}{lrrrrrrr}
\toprule
Dataset & Nodes & Edges & Edges & Reciprocity & Density & Clustering \\
        & $|V|$ & $|E|$ & (undirected) &  & $|E| / |V|^{2} $ & \\
 \midrule
Cora & 2,708 & 5,429 & 5,278 & 0.056 & 0.000741 & 0.131 \\
CiteSeer & 3,327 & 4,732 & 4,676 & 0.024 & 0.000428 & 0.074 \\
Google & 15,763 & 171,206 & 149,456 & 0.254 & 0.000689 & 0.343 \\
 \midrule
\end{tabular}
}
\end{table}

Table~\ref{datasets} gives the key network statistics. The Cora and CiteSeer graphs have few bidirectional edges. For the Google graph, however, a randomly sampled directed edge will be part of a bidirectional edge around 25\% of the time 

. Despite the difference in size, the graph density is similar across the datasets. The local density is higher for the Google dataset, as measured by the average (directed) clustering coefficient. Edge weights and/or node attributes are not considered. In all experiments, we use one-hot encoding of the node IDs as node features \cite{Salha2019}, except for the \textit{MAGNET} model for which we used in- and out-degree as prescribed by the authors \cite{zhang2021magnet}. While employing node IDs means the models are transductive, all the strategies can be extended to inductive settings by using other node features as appropriate.

\subsection{Models}

We evaluate our proposed training strategies using several NDLP-capable models from the literature, all of which follow the Graph Autoencoder paradigm. These include the Gravity-Inspired Graph Autoencoder (Gr-GAE)~\cite{Salha2019}, Source/Target Graph Autoencoder (ST-GAE)~\cite{Salha2019}, the DiGAE Directed Graph Auto-Encoder from~\cite{Kollias2022DirectedGA}, MAGNET~\cite{zhang2021magnet} and our custom MLP-GAE, which uses a decoder based on concatenating the encoder outputs followed by a multilayer perceptron. Each model is tested under various experimental conditions to evaluate its performance within our proposed framework. The standard graph autoencoder (GAE)~\cite{kipf2016variational} is included to provide an undirected baseline model. Further details on model implementations and settings are provided in Appendix~\ref{appendix-models}.

\subsection{Experimental Settings}

For the three tasks described in Fig.~\ref{fig:training_sets}, under the sampling defined in \S\ref{sec-split}, we measure ROC-AUC (Receiver Operating Characteristic - Area Under the Curve) to evaluate a model’s ability to distinguish between classes, while AUPRC (Area Under Precision-Recall Curve) evaluates precision across different recall levels. 
We train the models according to the strategies defined in \S\ref{sec-methods}, as well as a \textit{Baseline} strategy. Namely, the model is trained on the \textit{General DLP} training set with rebalancing of positive and negative edges' contributions to training loss (Binary Cross Entropy). For each novel training strategy, we perform early stopping on the sum of ROC-AUC and AUPRC metrics over the \textit{General DLP, Directional} and \textit{Bidirectional} validation sets. Missing self-loops are always inserted for message passing, and they are also used as positive supervision samples in both \textit{MO-NDLP} and \textit{S-NDLP}, while they are treated as bidirectional negative supervision samples in \textit{MC-NDLP}.

\begin{figure}[t!]
    \centering
    \includegraphics[width=0.9\linewidth]{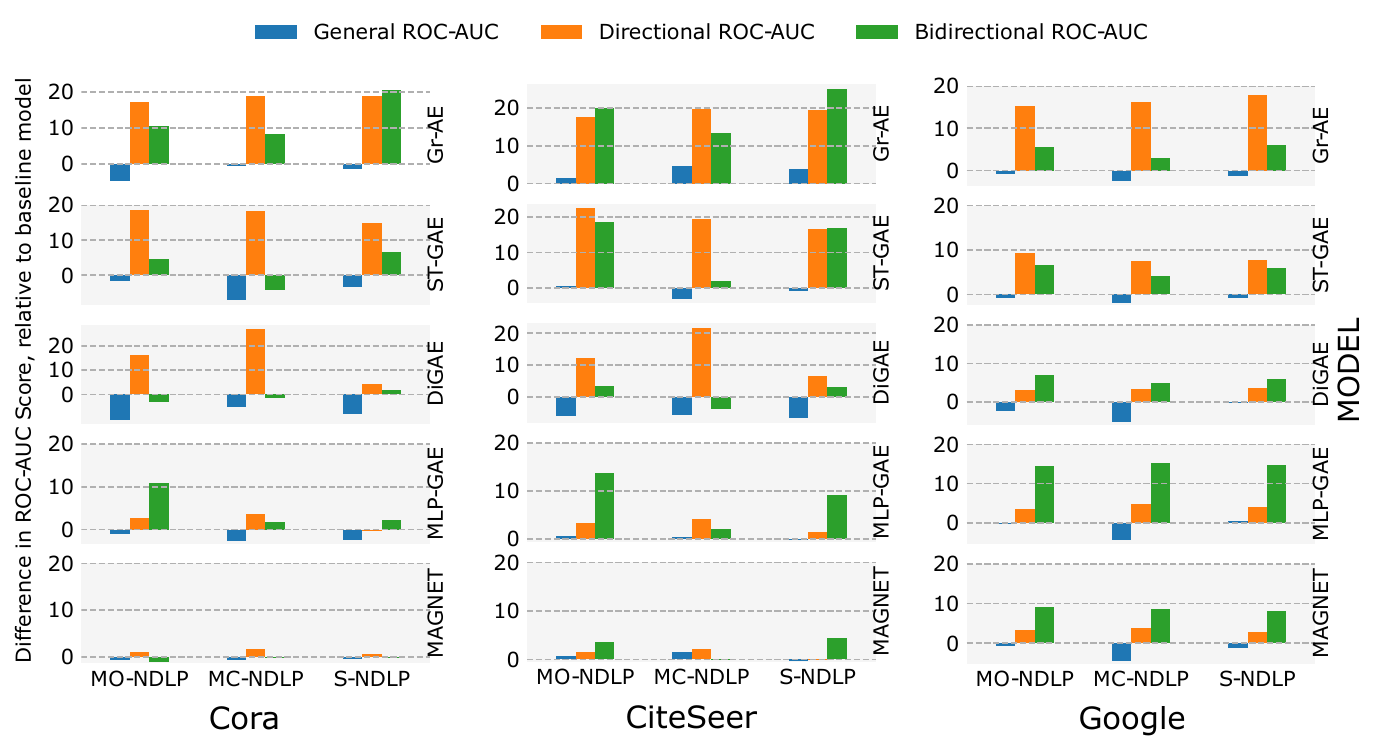}
    \caption{Performance difference of each proposed strategy compared to the baseline, measured in ROC-AUC  (x100). Each bar represents the change in ROC-AUC - either an increase or decrease - when applying one of the proposed strategies to a specific sub-task, NDLP model, and dataset, relative to the same model's baseline performance. Scores are averaged over 5 runs. Error bars are omitted for visual clarity.}
    \label{fig:results_barplots}
\end{figure}

\subsection{Results} \label{sec-results}

\begin{table*}[t!]
    \caption{ROC-AUC and AUPRC test scores of various models on Cora Dataset, trained with the \textit{Baseline} and our proposed strategies.}
    \label{results_table_cora}
    \begin{center}
    \begin{small}
    \begin{sc}
    \resizebox{0.85\linewidth}{!}{
    \begin{tabular}{@{}l@{~}l|l@{\quad}l|l@{\quad}l|l@{\quad}l@{}}
\toprule
\multicolumn{2}{}{} & \multicolumn{2}{c}{General} & \multicolumn{2}{c}{Directional} & \multicolumn{2}{c}{Bidirectional} \\
 &  & ROC-AUC & AUPRC & ROC-AUC & AUPRC & ROC-AUC & AUPRC \\
\midrule
\multirow[t]{4}{*}{GAE} & Baseline & 84.6 $\pm$ 0.4 & 88.6 $\pm$ 0.3 & 50.0 $\pm$ 0.0 & 50.0 $\pm$ 0.0 & 62.4 $\pm$ 3.0 & 64.0 $\pm$ 3.1 \\

\midrule
\multirow[t]{4}{*}{Gr-GAE} & Baseline & \underline{\textbf{89.2 $\pm$ 0.4}} & \underline{\textbf{92.4 $\pm$ 0.2}} & 63.4 $\pm$ 2.5 & 61.5 $\pm$ 2.7 & 69.1 $\pm$ 3.1 & 66.5 $\pm$ 3.3 \\
 & MO-NDLP & 84.5 $\pm$ 1.1 & 86.3 $\pm$ 1.1 & 80.6 $\pm$ 0.7 & 80.2 $\pm$ 0.9 & 79.6 $\pm$ 4.3 & 84.6 $\pm$ 3.5 \\
 & MC-NDLP & 88.6 $\pm$ 0.4 & 90.0 $\pm$ 0.4 & 82.1 $\pm$ 0.5 & \textbf{81.8 $\pm$ 0.7} & 77.3 $\pm$ 2.2 & 76.3 $\pm$ 1.7 \\
 & S-NDLP & 87.8 $\pm$ 0.6 & 89.5 $\pm$ 0.5 & \textbf{82.3 $\pm$ 0.5} & 81.6 $\pm$ 0.4 & \underline{\textbf{89.6 $\pm$ 1.6}} & \underline{\textbf{92.4 $\pm$ 1.1}} \\
\midrule
\multirow[t]{4}{*}{ST-GAE} & Baseline & \textbf{87.8 $\pm$ 0.7} & \textbf{90.1 $\pm$ 0.5} & 60.8 $\pm$ 0.5 & 64.5 $\pm$ 0.6 & 74.6 $\pm$ 1.8 & 74.1 $\pm$ 2.2 \\
 & MO-NDLP & 86.3 $\pm$ 0.5 & 86.2 $\pm$ 0.4 & \textbf{79.3 $\pm$ 1.0} & 80.0 $\pm$ 0.9 & 79.3 $\pm$ 0.5 & 79.5 $\pm$ 1.9 \\
 & MC-NDLP & 80.7 $\pm$ 2.0 & 80.1 $\pm$ 2.1 & 79.0 $\pm$ 2.3 & \textbf{81.6 $\pm$ 1.9} & 70.3 $\pm$ 3.0 & 68.1 $\pm$ 2.1 \\
 & S-NDLP & 84.5 $\pm$ 0.4 & 84.9 $\pm$ 0.7 & 75.8 $\pm$ 1.0 & 78.4 $\pm$ 0.9 & \textbf{81.1 $\pm$ 0.9} & \textbf{80.4 $\pm$ 1.6} \\
\midrule
\multirow[t]{4}{*}{DiGAE} & Baseline & \textbf{80.4 $\pm$ 1.1} & \textbf{85.3 $\pm$ 0.8} & 57.5 $\pm$ 1.3 & 63.0 $\pm$ 1.4 & 70.4 $\pm$ 2.2 & 68.6 $\pm$ 1.2 \\
 & MO-NDLP & 70.2 $\pm$ 3.8 & 72.6 $\pm$ 3.6 & 73.6 $\pm$ 5.4 & 76.0 $\pm$ 4.2 & 67.3 $\pm$ 4.6 & 69.6 $\pm$ 4.1 \\
 & MC-NDLP & 75.4 $\pm$ 0.9 & 77.4 $\pm$ 1.0 & \textbf{84.3 $\pm$ 0.6} & \textbf{85.4 $\pm$ 0.8} & 68.9 $\pm$ 1.5 & 69.3 $\pm$ 1.1 \\
 & S-NDLP & 72.5 $\pm$ 4.0 & 77.4 $\pm$ 4.4 & 61.6 $\pm$ 1.3 & 69.2 $\pm$ 1.4 & \textbf{72.1 $\pm$ 5.6} & \textbf{74.4 $\pm$ 5.7} \\
\midrule
\multirow[t]{4}{*}{MLP-GAE} & Baseline & \textbf{77.1 $\pm$ 0.9} & \textbf{78.2 $\pm$ 0.6} & 90.7 $\pm$ 0.6 & 90.7 $\pm$ 0.6 & 69.9 $\pm$ 3.2 & 69.7 $\pm$ 3.7 \\
 & MO-NDLP & 76.0 $\pm$ 0.8 & 76.4 $\pm$ 0.7 & 93.4 $\pm$ 0.6 & 93.5 $\pm$ 0.6 & \textbf{80.7 $\pm$ 1.6} & \textbf{79.2 $\pm$ 2.4} \\
 & MC-NDLP & 74.5 $\pm$ 0.7 & 75.6 $\pm$ 0.7 & \underline{\textbf{94.3 $\pm$ 0.6}} & \underline{\textbf{94.4 $\pm$ 0.5}} & 71.7 $\pm$ 2.4 & 65.7 $\pm$ 1.8 \\
 & S-NDLP & 74.7 $\pm$ 1.0 & 74.9 $\pm$ 0.9 & 90.5 $\pm$ 0.7 & 90.0 $\pm$ 0.9 & 72.0 $\pm$ 2.6 & 70.5 $\pm$ 2.9 \\
\midrule
\multirow[t]{4}{*}{MAGNET} & Baseline & \textbf{75.2 $\pm$ 1.4} & \textbf{77.8 $\pm$ 1.0} & 90.4 $\pm$ 0.9 & 89.8 $\pm$ 0.8 & \textbf{71.9 $\pm$ 2.3} & \textbf{70.4 $\pm$ 2.8} \\
 & MO-NDLP & 74.4 $\pm$ 1.4 & 77.4 $\pm$ 1.1 & 91.3 $\pm$ 1.0 & 90.9 $\pm$ 1.0 & 70.6 $\pm$ 2.7 & 68.6 $\pm$ 2.7 \\
 & MC-NDLP & 74.4 $\pm$ 1.0 & 77.4 $\pm$ 1.0 & \textbf{92.1 $\pm$ 0.7} & \textbf{91.6 $\pm$ 0.7} & 71.8 $\pm$ 2.6 & 70.0 $\pm$ 2.6 \\
 & S-NDLP & 74.6 $\pm$ 1.3 & 77.5 $\pm$ 1.1 & 91.0 $\pm$ 1.0 & 90.4 $\pm$ 1.0 & 71.8 $\pm$ 2.8 & 70.2 $\pm$ 2.9 \\
\bottomrule
\end{tabular}
    }
    \end{sc}
    \end{small}
    \end{center}
    \caption{ROC-AUC and AUPRC test scores of various models on CiteSeer Dataset, trained with the \textit{Baseline} and our proposed strategies.}
    \label{results_table_citeseer}
    \begin{center}
    \begin{small}
    \begin{sc}
    \resizebox{0.85\linewidth}{!}{
    \begin{tabular}{@{}l@{~}l|l@{\quad}l|l@{\quad}l|l@{\quad}l@{}}
\toprule
\multicolumn{2}{}{} & \multicolumn{2}{c}{General} & \multicolumn{2}{c}{Directional} & \multicolumn{2}{c}{Bidirectional} \\
 &  & ROC-AUC & AUPRC & ROC-AUC & AUPRC & ROC-AUC & AUPRC \\
\midrule
\multirow[t]{4}{*}{GAE} & Baseline & 78.6 $\pm$ 0.7 & 84.1 $\pm$ 0.6 & 50.0 $\pm$ 0.0 & 50.0 $\pm$ 0.0 & 56.2 $\pm$ 3.8 & 59.3 $\pm$ 1.9 \\

\midrule
\multirow[t]{4}{*}{Gr-GAE} & Baseline & 77.0 $\pm$ 0.7 & 84.3 $\pm$ 0.6 & 55.7 $\pm$ 2.3 & 58.2 $\pm$ 3.2 & 72.5 $\pm$ 3.7 & 71.3 $\pm$ 4.4 \\
 & MO-NDLP & 78.6 $\pm$ 0.8 & 82.6 $\pm$ 1.3 & 73.4 $\pm$ 1.6 & 76.8 $\pm$ 1.2 & 92.6 $\pm$ 2.1 & 94.6 $\pm$ 1.5 \\
 & MC-NDLP & \underline{\textbf{81.9 $\pm$ 0.8}} & \underline{\textbf{85.1 $\pm$ 0.5}} & \textbf{75.5 $\pm$ 0.7} & \textbf{78.9 $\pm$ 0.6} & 85.9 $\pm$ 2.8 & 85.1 $\pm$ 3.1 \\
 & S-NDLP & 80.8 $\pm$ 0.9 & 84.4 $\pm$ 0.7 & 75.2 $\pm$ 1.0 & 78.2 $\pm$ 0.9 & \underline{\textbf{97.5 $\pm$ 1.1}} & \underline{\textbf{98.0 $\pm$ 0.7}} \\
\midrule
\multirow[t]{4}{*}{ST-GAE} & Baseline & 80.9 $\pm$ 0.8 & \textbf{85.2 $\pm$ 0.7} & 56.0 $\pm$ 0.3 & 61.1 $\pm$ 0.5 & 72.0 $\pm$ 4.5 & 73.0 $\pm$ 3.7 \\
 & MO-NDLP & \textbf{81.4 $\pm$ 1.5} & 82.6 $\pm$ 2.0 & \textbf{78.5 $\pm$ 2.3} & \textbf{80.1 $\pm$ 1.8} & \textbf{90.5 $\pm$ 4.4} & \textbf{92.2 $\pm$ 4.1} \\
 & MC-NDLP & 77.8 $\pm$ 1.8 & 79.3 $\pm$ 2.4 & 75.5 $\pm$ 4.0 & 79.5 $\pm$ 2.8 & 73.9 $\pm$ 5.1 & 75.1 $\pm$ 4.9 \\
 & S-NDLP & 80.0 $\pm$ 1.3 & 82.0 $\pm$ 1.4 & 72.6 $\pm$ 1.6 & 77.4 $\pm$ 1.1 & 88.9 $\pm$ 4.3 & 90.3 $\pm$ 3.9 \\
\midrule
\multirow[t]{4}{*}{DiGAE} & Baseline & \textbf{78.5 $\pm$ 0.9} & \textbf{83.5 $\pm$ 0.8} & 56.6 $\pm$ 1.0 & 65.2 $\pm$ 1.5 & 62.3 $\pm$ 3.3 & 65.8 $\pm$ 3.8 \\
 & MO-NDLP & 72.6 $\pm$ 5.0 & 74.9 $\pm$ 5.2 & 68.7 $\pm$ 3.4 & 71.7 $\pm$ 4.1 & \textbf{65.6 $\pm$ 4.5} & 70.8 $\pm$ 5.2 \\
 & MC-NDLP & 72.7 $\pm$ 1.3 & 74.6 $\pm$ 0.9 & \textbf{78.3 $\pm$ 2.9} & \textbf{80.1 $\pm$ 1.8} & 58.6 $\pm$ 3.8 & 61.6 $\pm$ 3.3 \\
 & S-NDLP & 71.8 $\pm$ 3.9 & 75.4 $\pm$ 4.6 & 63.3 $\pm$ 1.4 & 69.7 $\pm$ 2.2 & 65.5 $\pm$ 5.1 & \textbf{71.5 $\pm$ 5.8} \\
\midrule
\multirow[t]{4}{*}{MLP-GAE} & Baseline & 73.3 $\pm$ 0.8 & \textbf{76.1 $\pm$ 0.7} & 88.4 $\pm$ 0.7 & 89.8 $\pm$ 0.6 & 76.5 $\pm$ 1.1 & 76.5 $\pm$ 2.6 \\
 & MO-NDLP & \textbf{74.0 $\pm$ 0.9} & 75.2 $\pm$ 1.0 & 91.8 $\pm$ 0.5 & 92.2 $\pm$ 0.5 & \textbf{90.2 $\pm$ 0.9} & \textbf{90.0 $\pm$ 1.4} \\
 & MC-NDLP & 73.7 $\pm$ 0.8 & 74.3 $\pm$ 0.9 & \underline{\textbf{92.6 $\pm$ 0.5}} & \underline{\textbf{92.9 $\pm$ 0.4}} & 78.5 $\pm$ 1.1 & 73.6 $\pm$ 2.4 \\
 & S-NDLP & 73.3 $\pm$ 0.7 & 74.8 $\pm$ 0.9 & 89.8 $\pm$ 0.3 & 90.1 $\pm$ 0.3 & 85.5 $\pm$ 2.5 & 85.1 $\pm$ 2.4 \\
\midrule
\multirow[t]{4}{*}{MAGNET} & Baseline & 71.6 $\pm$ 0.7 & 74.9 $\pm$ 0.8 & 89.5 $\pm$ 0.6 & 89.9 $\pm$ 0.6 & 70.9 $\pm$ 6.1 & 68.9 $\pm$ 6.6 \\
 & MO-NDLP & 72.3 $\pm$ 0.6 & 74.7 $\pm$ 0.6 & 91.0 $\pm$ 0.6 & 91.2 $\pm$ 0.5 & 74.6 $\pm$ 7.1 & 73.4 $\pm$ 7.7 \\
 & MC-NDLP & \textbf{73.2 $\pm$ 0.9} & \textbf{75.2 $\pm$ 0.9} & \textbf{91.6 $\pm$ 0.6} & \textbf{91.7 $\pm$ 0.6} & 71.1 $\pm$ 7.5 & 69.7 $\pm$ 7.5 \\
 & S-NDLP & 71.3 $\pm$ 0.8 & 74.6 $\pm$ 0.8 & 89.6 $\pm$ 0.6 & 90.0 $\pm$ 0.5 & \textbf{75.3 $\pm$ 6.2} & \textbf{73.5 $\pm$ 7.0} \\
\bottomrule
\end{tabular}

    }
    \end{sc}
    \end{small}
    \end{center}
\end{table*}

\begin{table*}[h!]
    \caption{ROC-AUC and AUPRC test scores of various models on Google Dataset, trained with the \textit{Baseline} and our proposed strategies.}
    \label{results_table_google}
    \begin{center}
    \begin{small}
    \begin{sc}
    \resizebox{0.85\linewidth}{!}{
    \begin{tabular}{@{}l@{~}l|l@{\quad}l|l@{\quad}l|l@{\quad}l@{}}
\toprule
\multicolumn{2}{}{} & \multicolumn{2}{c}{General} & \multicolumn{2}{c}{Directional} & \multicolumn{2}{c}{Bidirectional} \\
 &  & ROC-AUC & AUPRC & ROC-AUC & AUPRC & ROC-AUC & AUPRC \\
\midrule
\multirow[t]{3}{*}{GAE} & Baseline & 93.5 $\pm$ 0.2 & 94.9 $\pm$ 0.2 & 50.0 $\pm$ 0.0 & 50.0 $\pm$ 0.0 & 54.8 $\pm$ 0.8 & 53.6 $\pm$ 1.4 \\

\midrule
\multirow[t]{3}{*}{Gr-GAE} & Baseline & \textbf{98.3 $\pm$ 0.1} & \underline{\textbf{98.9 $\pm$ 0.0}} & 76.5 $\pm$ 0.8 & 69.1 $\pm$ 0.9 & 92.0 $\pm$ 0.2 & 91.9 $\pm$ 0.2 \\
 & MO-NDLP & 97.4 $\pm$ 0.1 & 98.1 $\pm$ 0.1 & 91.9 $\pm$ 0.2 & 90.3 $\pm$ 0.4 & 97.6 $\pm$ 0.1 & 97.6 $\pm$ 0.1 \\
 & MC-NDLP & 95.7 $\pm$ 0.1 & 95.7 $\pm$ 0.1 & 92.6 $\pm$ 0.2 & 92.8 $\pm$ 0.1 & 95.1 $\pm$ 0.1 & 94.2 $\pm$ 0.1 \\
 & S-NDLP & 96.9 $\pm$ 0.1 & 97.7 $\pm$ 0.0 & \textbf{94.3 $\pm$ 0.1} & \textbf{94.7 $\pm$ 0.1} & \textbf{98.0 $\pm$ 0.0} & \textbf{98.5 $\pm$ 0.0} \\
\midrule
\multirow[t]{4}{*}{ST-GAE} & Baseline & \underline{\textbf{98.4 $\pm$ 0.1}} & \textbf{98.7 $\pm$ 0.0} & 87.2 $\pm$ 0.2 & 86.2 $\pm$ 0.1 & 92.2 $\pm$ 0.3 & 89.6 $\pm$ 0.4 \\
 & MO-NDLP & 97.6 $\pm$ 0.2 & 97.4 $\pm$ 0.3 & \textbf{96.6 $\pm$ 0.1} & \textbf{96.8 $\pm$ 0.1} & \underline{\textbf{98.8 $\pm$ 0.1}} & \underline{\textbf{98.6 $\pm$ 0.1}} \\
 & MC-NDLP & 96.6 $\pm$ 0.1 & 96.4 $\pm$ 0.2 & 94.6 $\pm$ 0.1 & 96.1 $\pm$ 0.1 & 96.4 $\pm$ 0.1 & 95.9 $\pm$ 0.1 \\
 & S-NDLP & 97.6 $\pm$ 0.0 & 97.5 $\pm$ 0.1 & 95.0 $\pm$ 0.1 & 96.0 $\pm$ 0.1 & 98.3 $\pm$ 0.1 & 96.8 $\pm$ 0.1 \\
\midrule
\multirow[t]{4}{*}{DiGAE} & Baseline & \textbf{97.0 $\pm$ 0.1} & \textbf{97.8 $\pm$ 0.1} & 92.9 $\pm$ 0.2 & 94.5 $\pm$ 0.2 & 90.9 $\pm$ 0.3 & 87.7 $\pm$ 0.4 \\
 & MO-NDLP & 94.7 $\pm$ 0.1 & 95.7 $\pm$ 0.2 & 95.9 $\pm$ 0.1 & 96.9 $\pm$ 0.1 & \textbf{97.7 $\pm$ 0.1} & \textbf{97.9 $\pm$ 0.1} \\
 & MC-NDLP & 91.7 $\pm$ 0.6 & 92.4 $\pm$ 0.5 & 96.3 $\pm$ 0.2 & \textbf{97.2 $\pm$ 0.1} & 95.7 $\pm$ 0.3 & 95.5 $\pm$ 0.4 \\
 & S-NDLP & 96.8 $\pm$ 0.1 & 97.3 $\pm$ 0.1 & \textbf{96.5 $\pm$ 0.1} & 97.0 $\pm$ 0.1 & 96.7 $\pm$ 0.2 & 96.3 $\pm$ 0.3 \\
\midrule
\multirow[t]{3}{*}{MLP-GAE} & Baseline & 90.8 $\pm$ 0.1 & 91.6 $\pm$ 0.0 & 93.5 $\pm$ 0.1 & 94.4 $\pm$ 0.1 & 81.2 $\pm$ 0.2 & 77.8 $\pm$ 0.4 \\
 & MO-NDLP & 90.4 $\pm$ 0.1 & 91.0 $\pm$ 0.1 & 97.0 $\pm$ 0.0 & 97.3 $\pm$ 0.0 & 95.6 $\pm$ 0.1 & 95.1 $\pm$ 0.1 \\
 & MC-NDLP & 86.3 $\pm$ 0.1 & 87.9 $\pm$ 0.1 & \underline{\textbf{98.4 $\pm$ 0.1}} & \underline{\textbf{98.5 $\pm$ 0.1}} & \textbf{96.4 $\pm$ 0.2} & 95.4 $\pm$ 0.1 \\
 & S-NDLP & \textbf{91.2 $\pm$ 0.1} & \textbf{91.9 $\pm$ 0.1} & 97.6 $\pm$ 0.0 & 97.8 $\pm$ 0.0 & 96.0 $\pm$ 0.1 & \textbf{95.5 $\pm$ 0.1} \\
\midrule
\multirow[t]{3}{*}{MAGNET} & Baseline & \textbf{89.1 $\pm$ 0.1} & \textbf{90.1 $\pm$ 0.0} & 93.8 $\pm$ 0.6 & 94.3 $\pm$ 0.4 & 83.9 $\pm$ 2.0 & 77.7 $\pm$ 2.3 \\
 & MO-NDLP & 88.5 $\pm$ 0.2 & 89.8 $\pm$ 0.1 & 97.1 $\pm$ 0.1 & 97.2 $\pm$ 0.1 & \textbf{92.9 $\pm$ 0.1} & \textbf{91.1 $\pm$ 0.3} \\
 & MC-NDLP & 84.7 $\pm$ 0.7 & 86.2 $\pm$ 0.4 & \textbf{97.6 $\pm$ 0.0} & \textbf{97.3 $\pm$ 0.1} & 92.5 $\pm$ 0.2 & 88.9 $\pm$ 0.3 \\
 & S-NDLP & 87.9 $\pm$ 0.3 & 89.4 $\pm$ 0.2 & 96.7 $\pm$ 0.1 & 96.8 $\pm$ 0.1 & 91.9 $\pm$ 0.5 & 88.3 $\pm$ 0.9 \\
\bottomrule
\end{tabular}

    }
    \end{sc}
    \end{small}
    \end{center}
\end{table*}

The performance results are summarized in Table~\ref{results_table_cora} (Cora dataset), Table~\ref{results_table_citeseer} (CiteSeer dataset), Table~\ref{results_table_google} (Google dataset) and in Figure~\ref{fig:results_barplots}. Performances are averaged over 5 random splits, keeping the same seed for all models.
All ROC-AUC and AUPRC values are scaled by 100 for compactness and clearer visualization. In bold we highlight the best training strategy for each
metric/model/task combination, while the underlined scores indicate the best training strategy across
all models.

As a baseline, we evaluated an undirected graph autoencoder (GAE) model on the \textit{General DLP} NDLP sub-task. While GAE performs deceptively well on the \textit{General DLP} task, it fails to capture edge directionality, as expected. This is reflected in its random performance on the \textit{Directional} task, with a ROC-AUC score of 0.5. This limitation arises from the inner product decoder used by GAE~\cite{kipf2016variational}, which inherently assigns the same probability to both $(u,v)$ and $(v,u)$ edges. Results from this experiment are reported in the first rows of Tables~\ref{results_table_cora}, \ref{results_table_citeseer} and \ref{results_table_google}.

Our proposed strategies consistently improved performance on the \textit{Directional} and on the \textit{Bidirectional} tasks across all datasets and models, only slightly compromising (at times even benefiting) \textit{General DLP} performance \cite{zhang2021magnet}, with a few exceptions. For instance, \textit{MAGNET} showed similar performance on Cora and CiteSeer, regardless of the training strategy, while it achieved significant improvement in the Bidirectional task on the Google dataset when trained using our strategies. This highlights that even though some models, like \textit{MAGNET}, show limited gains on specific datasets, the overall benefits might be more pronounced in larger datasets like Google.

For other models like DiGAE, we observed a trade-off: its performance on the Directional task improved, but often at the expense of lower General task scores. Notably, on the Google dataset, especially with the S-NDLP strategy, DiGAE maintained its General task performance while delivering modest gains in Directional and Bidirectional tasks.
%
Both MLP-GAE and MAGNET performed well on the Directional task but struggled on the General task, where their scores were systematically lower than those of the baseline GAE. DiGAE also struggled with the General task, surpassing GAE’s baseline performance only on the Google dataset.

Selecting the right model-strategy combination depends on how much one is willing to sacrifice General task performance for improvements in Directional and Bidirectional tasks. Interestingly, this trade-off is not always necessary. For example, with the CiteSeer dataset, Gravity-GAE with MC-NDLP achieved the best General task performance while significantly improving Directional and Bidirectional scores.
However, the optimal combination of model and strategy varies by dataset. For the Cora dataset, Gravity-AE with S-NDLP offers a balanced solution, delivering strong Directional and Bidirectional performance with only a slight reduction in General task scores. On the CiteSeer dataset, ST-GAE with MO-NDLP provides a good balance, offering competitive General task performance alongside noticeable gains in Directional and Bidirectional tasks. Similarly, for the Google dataset, ST-GAE with MO-NDLP proves to be an excellent choice, delivering significant improvements in Directional and Bidirectional tasks with minimal sacrifice in General task performance.

\section{Conclusions} \label{sec-conclusions}

In this paper, we introduced and evaluated new training strategies to improve performance on Neural Directed Link Prediction tasks, addressing the limitations of current models in learning edge directionality. By extending existing models to handle multiple sub-tasks simultaneously, we demonstrated that the proposed strategies -- Multi-Class (MC-DLP), Scalarization-based (S-DLP), and Multi-Objective (MO-DLP) Directed Link Prediction -- consistently improve performance on both Directional and Bidirectional tasks, although at times with a trade-off in General task performance.

While no single approach universally outperforms across all settings, the flexibility offered by our proposed training strategies provides a powerful means for improving NDLP model capabilities.
Future work can focus on refining these strategies to minimize trade-offs, particularly for applications that demand robust handling of directed graphs and directed link prediction. Our training strategies for learning edge directionality might also be usefully combined with approaches that allow GNNs to better represent edge directionality. Many alternative encodings [4,5] and labeling tricks [1,2,3] have been proposed to enhance the expressiveness of GNNs, also for performing DLP, and it would be interesting to explore a wider range of augmented models. Simultaneous training across the three facets of DLP allows for more concise comparative studies to be done on the ability of models, and various enhancements, to provide balanced performance across the three facets of DLP.
Also, an interesting area for future exploration is knowledge graphs (KG), which could greatly benefit from our methods.
Since KG-oriented tasks often employ specialized losses with margin terms \cite{bordes2013translating} and involve complex query answering rather than basic link prediction \cite{ren2020query2box}, studying how enhanced directionality learning impacts KG performance would be a valuable direction.

\FloatBarrier
\bibliographystyle{unsrtnat}
\bibliography{reference}

\newpage

\appendix
\section{Appendix}

\subsection{Models}
\label{appendix-models}

We test our three frameworks on different NDLP models found throughout the literature. All the models we tried fall within the Graph Autoencoder paradigm. In the following, assuming $\vec{z}_v \in \mathbb{R}^L$ is the output encoder embedding for node $v$, $\hat{p}_{uv}$ is the predicted probability of edge $e_{uv}$ and $\sigma$ is the sigmoid function. We present a brief description for each of them.

Assuming $\vec{z}_v \in \mathbb{R}^L$ is the output encoder embedding for node $v$, $\hat{p}_{uv}$ is the predicted probability of edge $e_{uv}$ and $\sigma$ is the sigmoid function.

\subsubsection{GAE}

The encoder has 2 layers, with input dimension equal to the number of nodes of the graph it is trained on (due to OHE), hidden dimension 64 and output dimension 32. For more details, see \cite{kipf2016variational}. 

\subsubsection{Gravity-GAE}

The encoder is given by two layers of:

$$H^{l+1} = D^{-1}_{\text{out}} \tilde{A}XW$$

Where $\tilde{A} = A + I$ and $D^{-1}_{\text{out}}$ is the out-degree diagonal matrix of $\tilde{A}$. The input dimension i equal to the number of nodes (due to OHE), hidden dimension is 64 and output dimension 32. 

Its decoder is given by:

$$\hat{p}_{uv} = \sigma(\vec{z}_v[0] - \lambda\ln(||\vec{z}_u[1:] - \vec{z}_v[1:]||_2^2)$$

Taken from \cite{Salha2019}.

\subsubsection{Source/Target GAE}

Assuming $L$ is even, Its decoder is given by:

$$\hat{p}_{uv} = \sigma(\vec{z}_v[:\frac{L}{2}] \cdot \vec{z}_u[\frac{L}{2}:])$$

Taken from \cite{Salha2019}. Encoder and dimensions as Gravity-GAE.

\subsubsection{MLP-GAE}

In all frameworks except \textit{MC-NDLP}, its decoder is given by:

$$\hat{p}_{uv} = MLP(\vec{z}_v || \vec{z}_u)$$

Where $||$ represents concatenation. In the \textit{MC-NDLP} framework, it's defined as in \ref{sec-MCNDLP}. In both cases, the MLP has only one layer which allows for a much faster implementation (although specific to DLP). Encoder and dimensions as Gravity-GAE.

\subsubsection{DIGAE}

The encoder is given by:

\begin{equation}
\begin{cases}
\vec{z}^S_v &= \tilde{A}\sigma(\tilde{A}^T\vec{x}_vW^{(0)}_S)W^{(1)}_S\\
\vec{z}^T_v &= \tilde{A}^T\sigma(\tilde{A}\vec{x}_vW^{(0)}_T)W^{(1)}_T
\end{cases}
\end{equation}

And the decoder is the same as \textit{Source/Target-GAE}, where we recognize:

\begin{equation}
\begin{cases}
\vec{z}^S_v &= \vec{z}_v[:\frac{L}{2}]\\
\vec{z}^T_v &= \vec{z}_v[\frac{L}{2}:]
\end{cases}
\end{equation}

For further details, see \cite{Kollias2022DirectedGA}.

\subsubsection{MAGNET} 

The \textit{MAGNET} model is described in \cite{zhang2021magnet}. It consists of a GCN where the aggregation is performed via the Magnetic Laplacian, i.e. a complex hermitian matrix that transfers to directed graphs several useful properties once exclusively held by laplacians of undirected networks such has positive definitions.

\subsection{Hyperparameters}
\label{appendix-hyperparameters}

All models have two encoder layers. All have the following hyperparameters:

\begin{itemize}
\item \textit{input\_dimension}: always equal to the number of nodes in the graph, except for MAGNET where it is 2 (in- and out-degree);
\item \textit{hidden dimension}: equal to 64 for all models across datasets except MAGNET which has 16;
\item \textit{output dimension}: set to 32 for all models across datasets  except MAGNET that is configured as in \cite{zhang2021magnet} with few modifications (see below);
\item \textit{number of epochs}: set to 1000 for all models across datasets except MAGNET which has 3000.
\end{itemize}

The $\lambda$ parameter of Gravity-GAE is initially set across datasets as in \cite{Salha2019}, but it is subsequently trained together with all other model parameters. Notably, we started it from $\lambda=0.1$ when training for Multi-Class on Google for convergence reasons.

The $\alpha$ and $\beta$ parameters of DiGAE have been set to 0.5 and not trained due to occasional instability.

MLP-GAE and MAGNET have a custom one-layer MLP decoder specifically designed to compute the probability of each edge given its vertices' embeddings in an efficient way. It represents the same functions as a standard MLP, but its internal matrix operations have been reordered specifically for DLP. It has 0.5 dropout.

MAGNET's $q$ parameter has been set to $0.05$ and not trained to isolate our frameworks' effects, being it directly linked to directionality embedding \cite{zhang2021magnet}.

MAGNET's implementation has been taken from Pytorch Geometric Signed Directed \cite{he2024pytorch} and slightly modified to accomodate the custom MLP decoder. The order of the Chebyshev polynomial is 2 everywhere.

All models make use of early stopping. The validation metric is the sum of ROC-AUC and AP-AUC on the \textit{General} validation set and on the \textit{General, Directional and Bidirectional} validation sets respectively for the Baseline and all the other training frameworks. patience is set to 200 for all models except for MagNet under the Multi-Class framework where it was set to 30 to keep computational time tractable.

All missing self-loops were always added as positive training targets for the Baseline, Multi-Objective and Scalarization frameworks, while they were treated as negatives in the Multi-Class framework according to empirical trials and for standardization.

Adam optimizer was used to train all models \cite{kingma2017adammethodstochasticoptimization}, that were implemented in PyTorch. Learning rates are summarized in \ref{lr-table}  except for GAE that has $lr=0.05$ for all Baseline trainings. Weight decay has been set to 5e-4 for MAGNET only. 

All models' hyperparameters values have either been taken from the respective papers or empirically tuned to achieve stable learning across novel techniques.

\begin{table}[ht]
\caption{Models' learning rates}
\begin{center}
\begin{tabular}{c|c|c|c}
\toprule
    Dataset & Method & Model & lr\\
    \midrule
    \multirow{10}{*}{Cora} 
                                &   & GR-AE & 0.01\\ 
                                & Baseline,   & ST-GAE & 0.01\\ 
                                & Multi-Objective \& & MLP-GAE & 0.002\\ 
                                & Scalarization  & DiGAE & 0.02\\ 
                                &   & MAGNET & 0.001\\ \cline{2-4}
                & \multirow{5}{*}{Multi-Class}  
                                    & GR-AE & 0.01\\  
                                &   & ST-GAE & 0.01\\ 
                                &   & MLP-GAE & 0.001\\ 
                                &   & DiGAE & 0.02\\  
                                &   & MAGNET & 0.001\\  
    \midrule
    \multirow{10}{*}{Citeseer} 
                                &   & GR-AE & 0.05\\  
                                & Baseline,  & ST-GAE & 0.02\\  
                                & Multi-Objective \&  & MLP-GAE & 0.002\\  
                                & Scalarization  & DiGAE & 0.02\\ 
                                &   & MAGNET & 0.001\\ \cline{2-4}
                & \multirow{5}{*}{Multi-Class}  
                                    & GR-AE & 0.01\\ 
                                &   & ST-GAE & 0.01\\ 
                                &   & MLP-GAE & 0.001\\ 
                                &   & DiGAE & 0.02\\  
                                &   & MAGNET & 0.001\\ 
    \midrule
    \multirow{10}{*}{Citeseer} 
                                &   & GR-AE & 0.05\\ 
                                & Baseline,  & ST-GAE & 0.01\\ 
                                & Multi-Objective \&  & MLP-GAE & 0.002\\ 
                                & Scalarization  & DiGAE & 0.02\\ 
                                &   & MAGNET & 0.001\\ \cline{2-4}
                & \multirow{5}{*}{Multi-Class}  
                                    & GR-AE & 0.01\\ 
                                &   & ST-GAE & 0.01\\ 
                                &   & MLP-GAE & 0.002\\ 
                                &   & DiGAE & 0.002\\ 
                                &   & MAGNET & 0.001\\ 
    \bottomrule
\end{tabular}
\end{center}
\label{lr-table}
\end{table}

\subsection{Statistical Independence in the Multi-Class approach} \label{sec:stat-indep}

In order to prove Equation \ref{eq-multiclass_extension}, it suffices to prove that probabilities properly factorize:

\begin{equation}
\hat{p}^{pb}_{uv} = \hat{P}(e_{uv}, e_{vu} | A,X) = \hat{P}(e_{uv}| e_{vu}, A,X) \hat{P}(e_{vu} | A,X) = \hat{P}(e_{uv}| A,X) \hat{P}(e_{vu} | A,X) 
\end{equation}

And similarly for the other cases. We therefore proved that joint probabilities can be factorized and therefore Equation \ref{eq-multiclass_extension} applies. 

\end{document}